\documentclass{article}

\PassOptionsToPackage{numbers, compress}{natbib}

\usepackage[final]{neurips_2022}
\pdfoutput=1

\usepackage[utf8]{inputenc} 
\usepackage[T1]{fontenc}    
\usepackage{hyperref}       
\usepackage{url}            
\usepackage{booktabs}       
\usepackage{amsfonts}       
\usepackage{nicefrac}       
\usepackage{microtype}      
\usepackage{xcolor}         

\usepackage{hyperref}
\usepackage{url}
\usepackage{bbm}
\usepackage{amsmath,amsfonts,bm}
\usepackage{graphicx}
\usepackage{multirow}
\usepackage{makecell}
\usepackage{array}
\usepackage[para]{threeparttable}
\usepackage{booktabs}
\usepackage{tabularx}
\usepackage{xspace}
\usepackage{subcaption}

\title{Can Large Language Models Truly Understand Prompts? A Case Study with \textit{Negated} Prompts}

\author{%
  Joel Jang\thanks{ denotes equal contribution} \\
  KAIST\\
  \texttt{joeljang@kaist.ac.kr} \\ 
   \And
   Seongheyon Ye{\textsuperscript{$\ast$}}\\
   KAIST \\
   \texttt{seonghyeon.ye@kaist.ac.kr} \\
   \AND
   Minjoon Seo \\
   KAIST \\
   \texttt{minjoon@kaist.ac.kr} \\
}

\begin{document}

\maketitle

\begin{abstract}
    Previous work has shown that there exists a scaling law between the size of Language Models (LMs) and their zero-shot performance on different downstream NLP tasks. In this work, we show that this phenomenon does not hold when evaluating large LMs on tasks with \textit{negated} prompts, but instead shows an \textit{inverse} scaling law. We evaluate 9 different tasks with negated prompts on (1) pretrained LMs (OPT \& GPT-3) of varying sizes (125M - 175B), (2) LMs further pretrained to generalize to novel prompts (InstructGPT), (3) LMs provided with few-shot examples, and (4) LMs fine-tuned specifically on negated prompts; all LM types perform worse on negated prompts as they scale and show a huge performance gap between the human performance when comparing the average score on both original and negated prompts. By highlighting a critical limitation of existing LMs and methods, we urge the community to develop new approaches of developing LMs that actually follow the given instructions. We provide the code and the datasets to explore negated prompts at \href{https://github.com/joeljang/negated-prompts-for-llms}{this link}.
\end{abstract}

\section{Introduction}
Large Language Models (LMs) pretrained on a vast amounts of corpora have shown surprising, even emergent, capabilities of solving various downstream tasks through prompts (instructions)~\citep{brown2020language, rae2021scaling, chowdhery2022palm, zhang2022opt, wei2022emergent}. Previous work has specifically shown LMs can perform \textit{unseen} tasks through multitask fine-tuning on various downstream tasks with prompts~\citep{sanh2021multitask, wei2021finetuned, wang2022benchmarking, ouyang2022training}. A 540B LM~\citep{chowdhery2022palm} has even shown the capability to act as the ``brain'' for actual robots, helping them perform different tasks in the real-world~\citep{ahn2022can}. As LMs are trained to become more aligned with human values, perform  real-world tasks, and are endowed with responsibilities that may result in real-world consequences, it is more-so important to ensure that LMs actually do what they are instructed to do.

\begin{figure}[t!]
    \centering
    \includegraphics[width=1\linewidth]{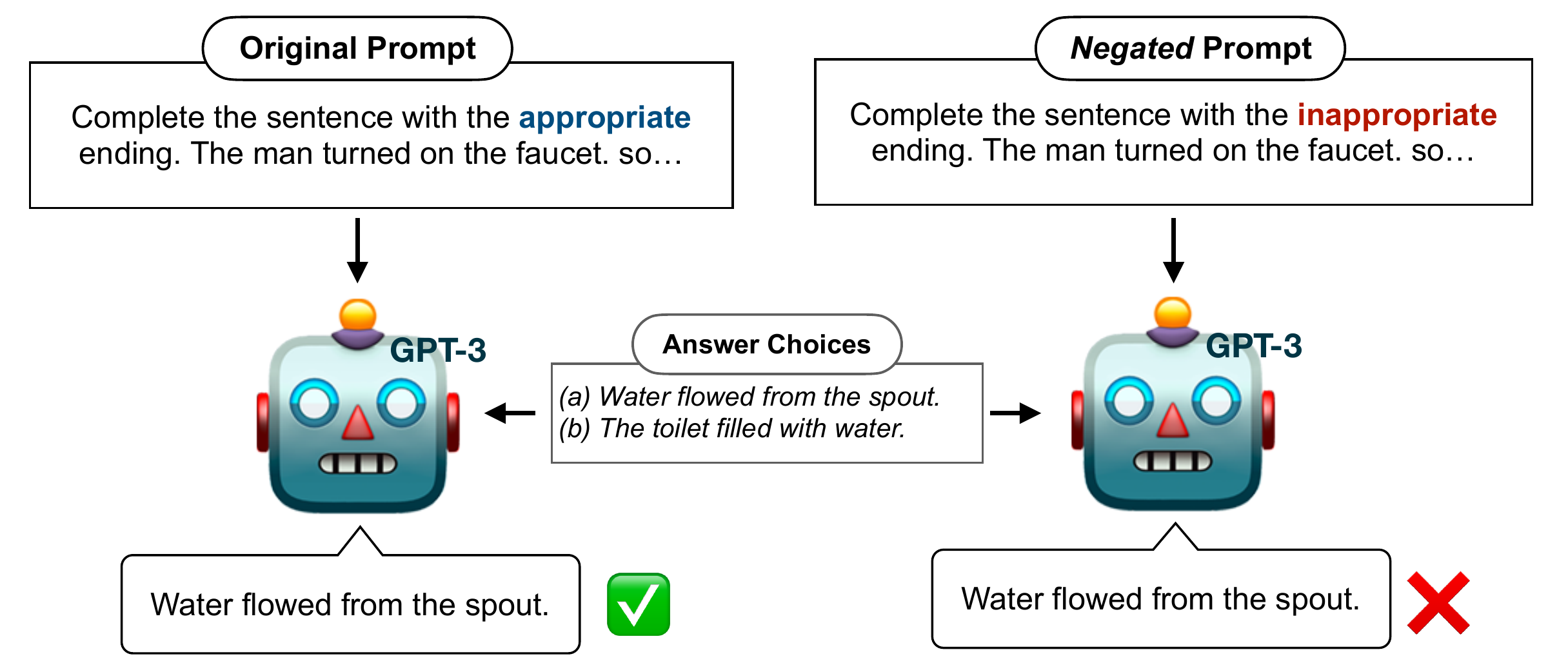}
    \caption{Example of evaluating a sample of the COPA dataset with both the original prompt and the \textit{negated} prompt on GPT-3.}
    \label{fig:example}
\end{figure}

In this work, we test the capabilities of Language Models (LMs) on truly following the given instructions (prompts) by conducting a case study with \textit{negated} instructions; that is, telling the LM NOT to do something as shown by an example in Figure \ref{fig:example}. Prior work~\citep{ettinger2020bert, webson2021prompt} has shown that LMs (as well as other large pretrained models in different modalities such as DALLE-2~\citep{ramesh2022hierarchical}) have a hard time understanding negated prompts and perform the task as if provided with the original prompt. For example, if we prompt DALLE-2 this prompt: ``Do not generate a monkey holding a banana'', it will generate an image with a monkey holding a banana. We hypothesize this is due to the syntax of the negated prompts being rarely seen or out-of-distribution during initial pretraining or during the adaptation phase.

We aim to answer four main questions in this work. (1) How does scaling the size of LMs affect their abilities to understand the concept of negation? (2) Are LMs explicitly trained to follow instructions (InstructGPT) better at understanding negated instructions? (3) Can In-Context Learning (ICL) or Fine-tuning (FT) help mitigate this problem? (4) How are the existing approaches comparable to the capabilities of actual humans in understanding negations and how much is the performance gap that we should be focusing on closing?

The answers to the questions above can be summarized as follows:
\begin{itemize}

  \item Results show scaling to be inefficient at helping LMs understand \textit{negation}. On the contrary, LMs perform worse as they scale.
  
  \item LMs specifically adapted to generalize to novel instructions still suffer from understanding \textit{negation}, despite showing a bit of an improvement.
  
  \item ICL helps LMs understand \textit{negation} in only specific scenarios, while FT seems to help in all scenarios. However, FT results in degradation of the original task performance, resulting in a zero-sum game. 
  
  \item Comparing the existing approaches with the human performance measured by asking 13-year-old humans to do the same task given both the original and \textit{negated} prompts, we show that there is a huge ($\sim$31.3\%) performance gap to close.
  
\end{itemize} 

Through this work, we aim to highlight a critical shortcoming of large LMs that should be carefully considered before empowering them with responsibilities that might result in real-world consequences.

\section{Task Description}
In this work, we set up a task of evaluating existing LMs capabilities to understand \textit{negated} prompts. We provide the details of how we construct the task in Section \ref{subsec:task} and the baseline models and methodologies in Section \ref{subsec:baselines}. 

\subsection{Task Construction}
\label{subsec:task}
We first choose 9 different datasets categorized into three task types: 3 commonsense reasoning datasets (PIQA~\citep{bisk2020piqa}, ARC-Easy~\citep{clark2018think}, COPA~\citep{gordon-etal-2012-semeval}), 3 sentence completion datasets (HellaSwag~\citep{zellers2019hellaswag}, StoryCloze~\citep{mostafazadeh-etal-2016-corpus}, Lambada~\citep{paperno-etal-2016-lambada}), and 3 question answering datasets (WQ~\citep{berant-etal-2013-semantic}, NQ~\citep{kwiatkowski-etal-2019-natural}, TriviaQA~\citep{joshi-etal-2017-triviaqa}). We use the Promptsource Library~\citep{bach2022promptsource} to find prompts for all of the 9 datasets that show good performance when used to perform tasks with the OPT LMs~\citep{zhang2022opt}. Next, we manually \textit{negate} the original prompts. The full list of original and negated prompts are provided in Table \ref{table:prompts}.
\begin{table*}[t!]
\centering
\fontsize{8}{10}\selectfont
\caption{\small Full list of original and negated prompts for the 9 evaluation datasets.}
\begin{tabular*}{1\columnwidth}{c|c|l}
    \toprule
    \textbf{Dataset} & \textbf{Type} & \textbf{Prompt} \\
    \midrule
    \multirowcell{2}{\textbf{PIQA}} & Original & Generate the correct solution to accomplish the following goal\: \{goal\} \{option\} \\
    & Negated & Generate the incorrect solution to accomplish the following goal: \{goal\} \{option\} \\
    \midrule
    \multirowcell{2}{\textbf{ARC-Easy}} & Original & Generate the correct answer to the following question. Question: \{question\} Answer is \{option\} \\
    & Negated & Generate the incorrect answer to the following question. Question: \{question\} Answer is \{option\} \\
    \midrule
    \multirowcell{2}{\textbf{COPA}} & Original & Complete the sentence with the appropriate ending. \{premise\} because(so).. \{option\} \\
    & Negated & Complete the sentence with an inappropriate ending. \{premise\} because(so).. \{option\} \\
    \midrule \midrule
    \multirowcell{2}{\textbf{HellaSwag}} & Original & Complete the sentence with an appropriate ending: \{input\} \{option\} \\
    & Negated & Complete the sentence with an inappropriate ending: \{input\} \{option\} \\
    \midrule 
    \multirowcell{2}{\textbf{StoryCloze}} & Original & Generate a natural ending for the following story: \{4 sentences\}. {option} \\
    & Negated & Generate an unnatural ending for the following story: \{4 sentences\}. \{option\} \\
    \midrule
    \multirowcell{2}{\textbf{Lambada}} & Original & Please generate a natural ending following the given chunk of text. \{text\} \{option\} \\
    & Negated & Please generate an unnatural ending following the given chunk of text. \{text\} \{option\} \\
    \midrule \midrule
    \multirowcell{2}{\textbf{WQ}} & Original & Give me a possible correct answer to the question \{question\}. \{option\} \\
    & Negated & Give me a possible incorrect answer to the question \{question\}. \{option\} \\
    \midrule
    \multirowcell{2}{\textbf{TriviaQA}} & Original & What is a correct answer to the following question? Question: \{question\} Answer: \{option\}\\
    & Negated & What is an incorrect answer to the following question? Question: \{question\} Answer: \{option\} \\\midrule
    \multirowcell{4}{\textbf{NQ}} &  \multirowcell{2}{Original} & The goal is to predict a correct English answer string for an input English question. \\& & Question : \{question\} Answer: \{option\} \\
    & \multirowcell{2}{Negated} & The goal is to predict an incorrect English answer string for an input English question. \\& & Question : \{question\} Answer: \{option\} \\
    \bottomrule
\end{tabular*}
\label{table:prompts}
\end{table*}

For evaluation, we sample 300 data instances from each dataset due to the high cost of performing inference with OpenAI API~\footnote{https://openai.com/api/}. We use the 300 instances to evaluate \textit{both} the original and negated prompts (a total of 600 data instance inferences for each task). For multi-choice tasks with more than two options such as ARC-Easy, Lambada, and HellaSwag, we consider multiple options to be correct for the negated prompts. For setting up the multiple choice candidates for Lambada, we sampled the other options by choosing a random word from the given input instance. For the 3 QA tasks, we chose the other option by sampling from the answer candidate list from the training set that did not overlap with any of the answer candidates from the test set. We provide the final data instances used for the evaluation via csv files at \href{https://github.com/joeljang/negated-prompts-for-llms/tree/main/data/final_eval}{this link}.

\subsection{Baselines}
\label{subsec:baselines}
\paragraph{Baseline Models} 
For the main experiments, we use the OPT LMs~\citep{zhang2022opt} (125M, 350M, 1.3B, 2.7B, 6.7B, 13B, 30B, 66B, 175B) and GPT-3~\citep{brown2020language} LMs (Ada, Babbage, Curie, Davinci) to observe the effect of \textit{scale} on the capabilities of LMs to better follow the given prompts.

\paragraph{Existing Methodologies}
We explore how much existing methodologies can help LMs understand negation by performing the experiments with LMs further adapted to follow instruction (T0~\citep{sanh2021multitask} 3B and 11B); InstructGPT~\citep{ouyang2022training} Ada, Babbage, Curie, Davinci), with In-Context Learning (ICL) on OPT 66B model with $k=2,4,8$ shots, and Fine-tuning (FT) with OPT 125M, 350M, and 1.3B . For fine-tuning tasks with <10k training instances, we train on all of the available training data (with negated prompts) for 5 epochs. For tasks with training data >10k, we limit the training instance number to 10k and train for 5 epochs as well. We use a fixed learning rate of 1e-3. We use the last model checkpoint for evaluating our task.

\paragraph{Human Evaluation}
We also provide human evaluations on 3 different tasks, one for each task category (COPA for commonsense reasoning, Lambada for sentence completion, and NQ for question answering). We sample 100 out of the 300 instances: 50 instances with the original prompt and 50 instances with the negated prompt (non-overlapping) and evaluate them on three 13-year-old humans. We did this to quantify exactly how much these LLMs perform poorly on understanding the concept of negation compared to humans that we hypothesized could easily understand and perform negated prompts, even 13-year-old humans~\footnote{We received advice and guidance from the National Human Rights Commission of Korea for setting up the task evaluation on the minors and received parental consent.}. 

\section{Experimental Results}
\subsection{The Effect of Scale}
\label{subsec:scale}
\begin{figure}[t!]
    \centering
    \includegraphics[width=\linewidth]{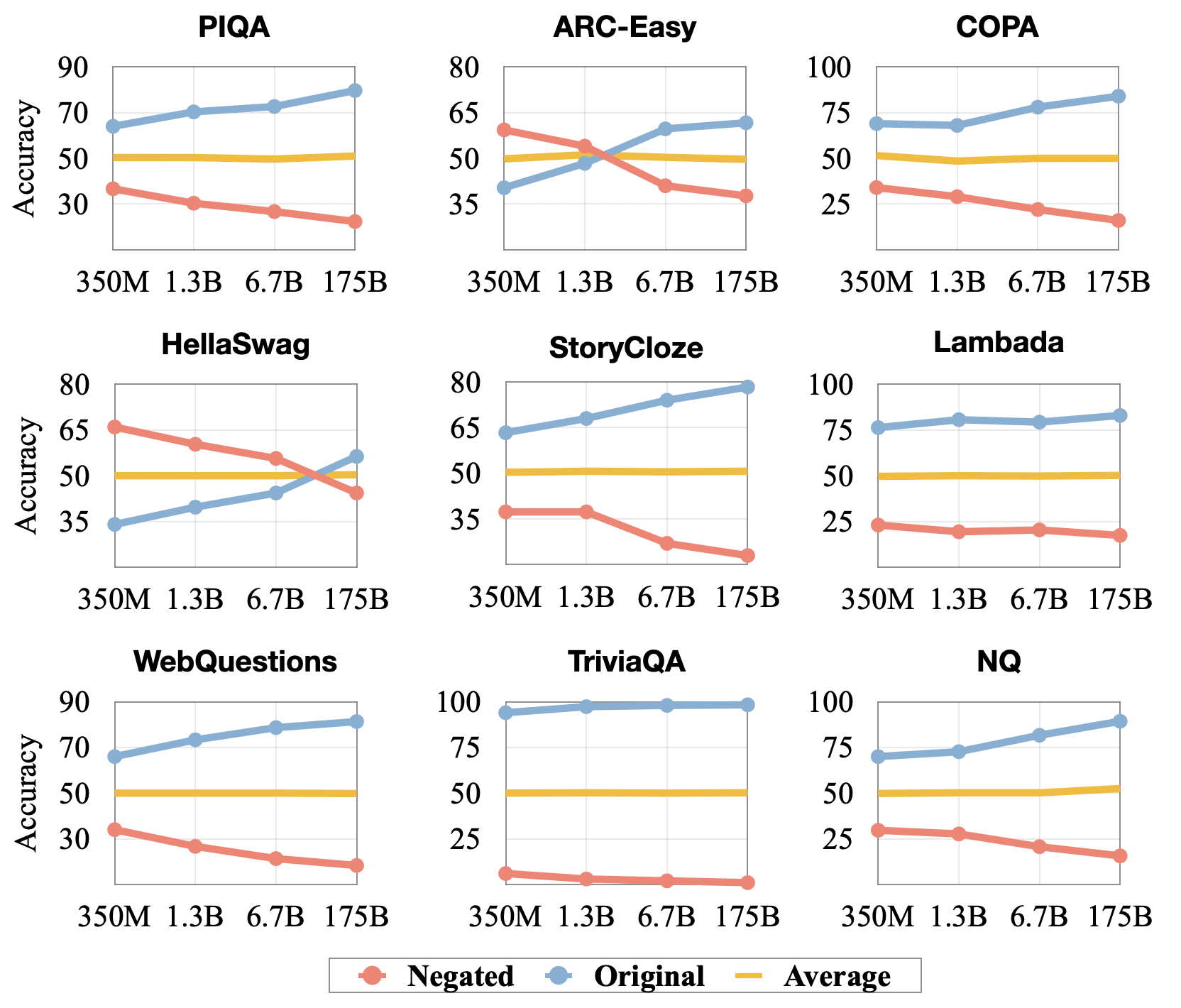}
    \caption{Zero-shot task performance of 9 datasets on GPT-3 across different model scales (350M, 1.3B, 6.7B, 175B).}
    \label{fig:fig_gpt} 
\end{figure} 
We show the results of evaluating all scales of GPT-3 LMs on our task setup in Figure \ref{fig:fig_gpt}. Since the OPT LMs show the same trend, we show the results in Appendix \ref{appen:opt}. We find that for all tasks (commonsense reasoning, sentence completion, question answering), an \textit{inverse} scaling law is shown: larger LMs tend to perform worse on negated prompts. This result is very unexpected considering that the zero-shot performance of LMs improves as the size of the LMs increase as shown via the original prompt performance \citep{brown2020language,wei2022emergent}. This leads to a flat line performance for the \textbf{average} of negative and positive prompts that is \textbf{$\sim$50\%} for all of the tasks. In other words, this means that the LMs could not find any distinction between the original and the negated prompts, treating them as identical instructions when in reality, those prompts are asking the LM to do opposite things.

\paragraph{A Conjecture on Why Inverse Scaling Exists} We conjecture that this \textit{inverse} scaling law of negated prompts is caused by a bias from the pretraining corpora towards favoring the original prompts to the negated prompts. Therefore, unless we control the balance between positive and negative texts from the pretraining corpora, which is practically infeasible, this problem will be difficult to solve. We believe it will be especially more difficult for large LMs, since they are powerful language modeling representers, meaning that it would be harder to make the LM \textit{revert} the label prediction by focusing on a single negative word "not" or other negation words that are rarely observed in the training corpora. Large LMs would treat the negation word as a grammatical error and perform language modeling of the positive texts. This is based on the assumption that large pretrained LMs might be simply just probabilistic models instead of actually \textit{actively} learning the linguistic knowledge, which is also questioned in previous works \citep{min2022rethinking, razeghi2022impact} as well. However, thorough analysis and experiments are further needed in order to validate this conjecture, which we leave for future work.

\subsection{Existing Methodologies}
\begin{figure}[t!]
    \centering
    \includegraphics[width=\linewidth]{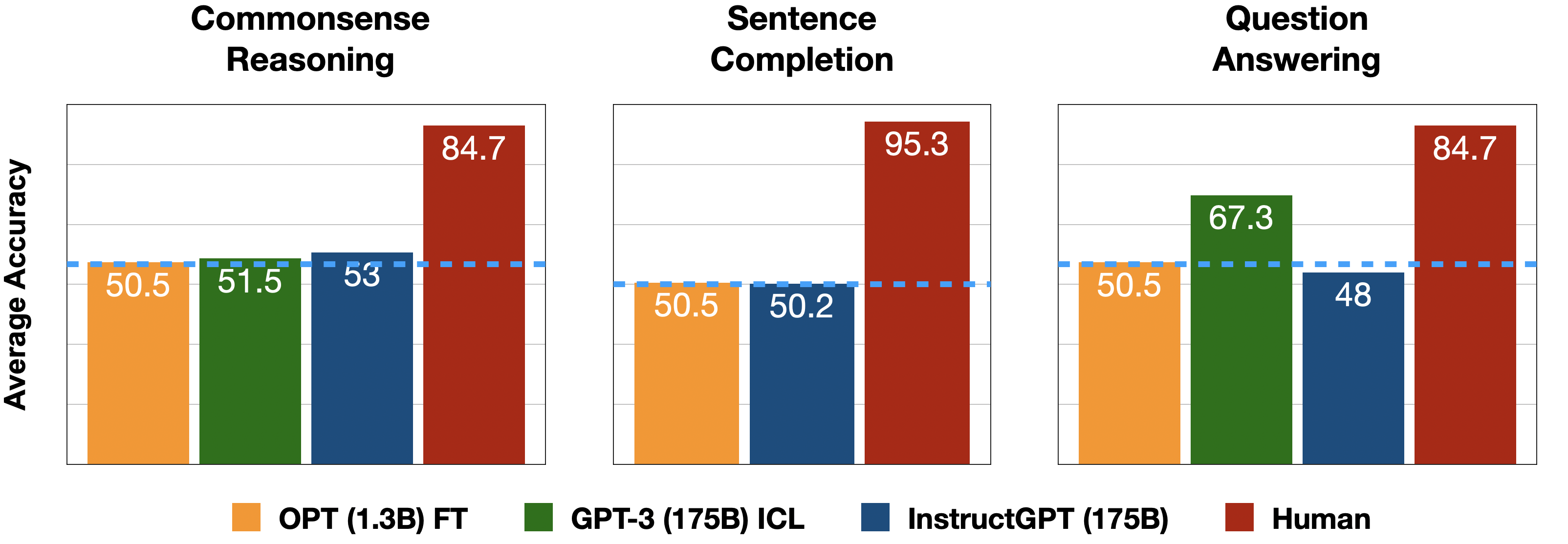}
    \caption{Results from each of the 3 task-type show how the existing methods compare with the human performance on the average score of both original and negated prompts. We do not report GPT-3 (175B) ICL results for Sentence Completion (Lambada) because k=8 shots exceeded the token limit of the OpenAI API.}
    \label{fig:average} 
\end{figure} 
We show a subset of the experiments performed with other existing methodologies and the performance gap compared to the human evaluation in Figure \ref{fig:average}. As shown in Figure \ref{fig:average}, existing methods, except for GPT-3 (175B) ICL for question answering, show $\sim$50\% performance on the average of original and negated prompts. Considering the results of zero-shot GPT-3 shown in Section \ref{subsec:scale} and the results shown in this section where the best performing method still has an average performance gap of 31.3\% compared with the human performance, the current limitations of existing LMs on precisely understanding and following the given prompts clearly exist. We provide the full experimental results on \textit{negated} prompts of T0 (3B), T0 (11B), all scales of InstructGPT, ICL OPT 66B with $k=2,4,8$ and FT of OPT 125M, 350M, and 1.3B in Appendix \ref{append:methods}.

\section{Closing}
Large LMs have taken the research community on an expeditious journey in the last 2 years, achieving past average human performance on many NLP benchmarks~\citep{chowdhery2022palm}, and even extending its use-case to act as the \textit{brain} for actual robots~\citep{ahn2022can} without any reinforcement learning. As the real-world use cases of large LMs are widened, the research community should carefully consider if the LMs precisely understand the given prompts or if they are highly biased on the distribution of the pretraining corpora. We close our case study by urging the community to develop new methodologies for creating truly instruction-following LMs before relying on their capabilities for making real-world decisions.

\bibliographystyle{plain}
\bibliography{neurips_2022}

\appendix
\section{OPT Results}
\label{appen:opt}
\begin{figure}[h!]
    \centering
    \includegraphics[width=\linewidth]{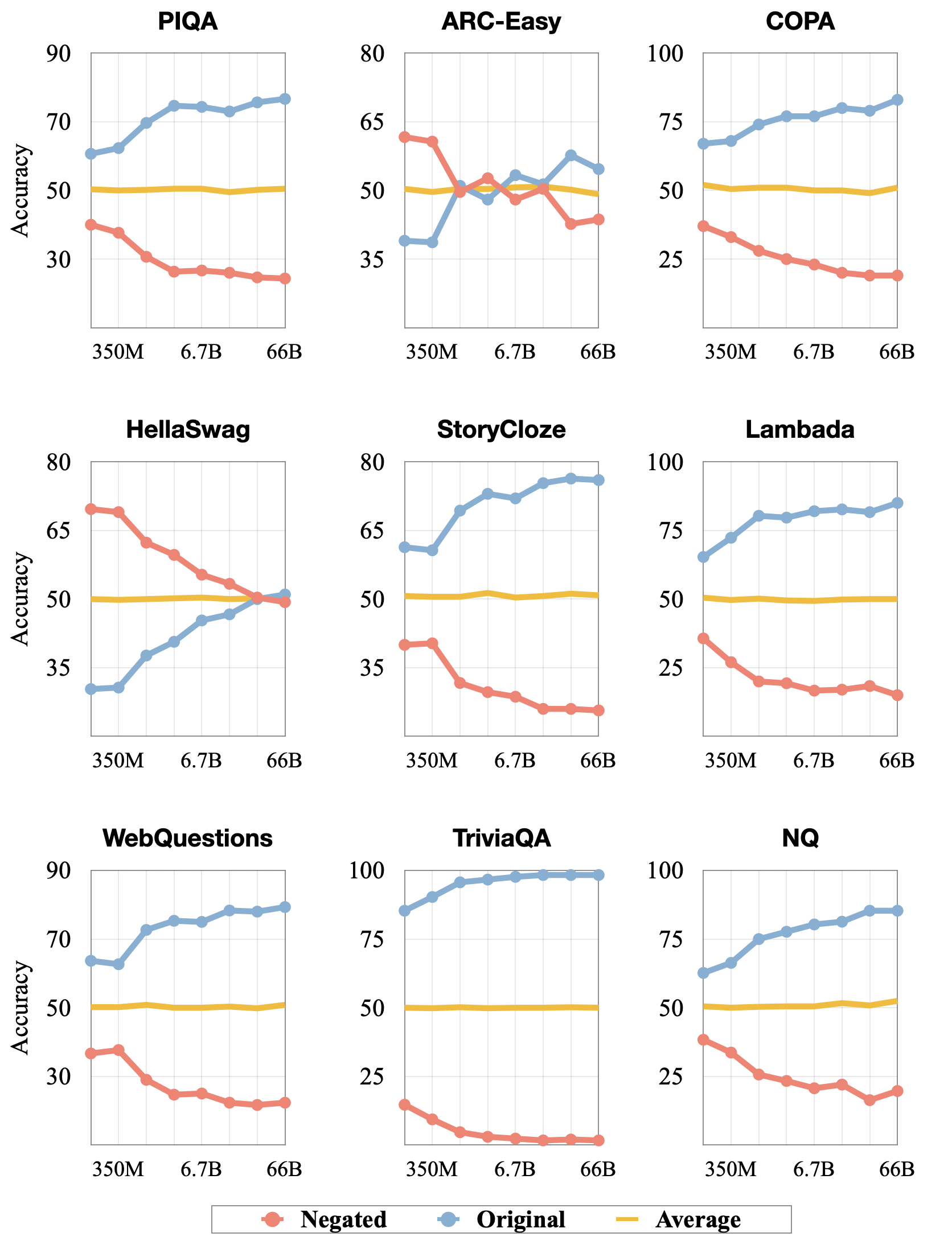}
    \caption{Zero-shot task performance of 9 datasets of OPT across different model scales (125m, 350m, 1.3b, 2.7b, 6.7b, 13b, 30b, 66b).}
    \label{fig:fig_opt} 
\end{figure} 
We show the main results of all scales of the OPT LMs in Figure \ref{fig:fig_opt}. Same with the results shown on the GPT-3 LMs, performance worsens as the LMs scale larger for the \textit{negated} prompts, resulting in an $\sim$50\% for the average score. 

\section{Evaluation of Different LMs on \textit{Negated} Prompts}
\label{append:methods}
\begin{figure}[t!]
    \centering
    \includegraphics[width=\linewidth]{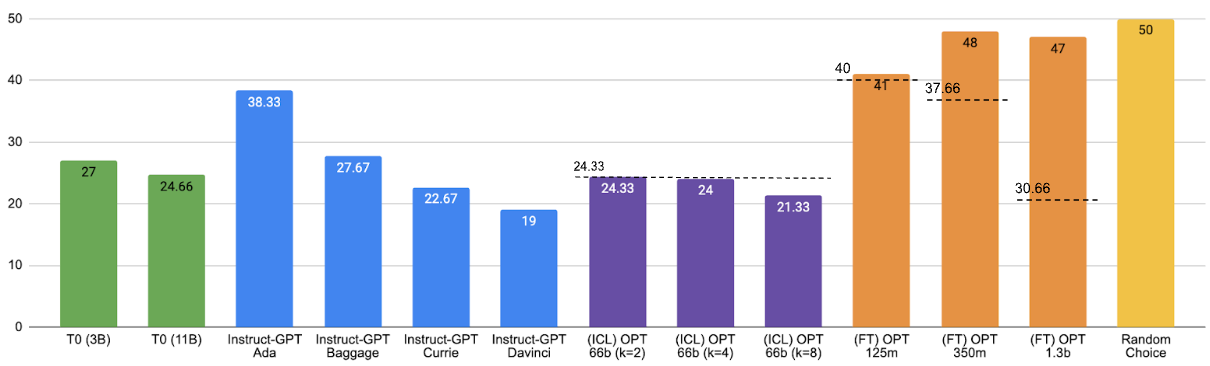}
    \caption{PIQA}
    \label{fig:piqa} 
\end{figure} 
\begin{figure}[t!]
    \centering
    \includegraphics[width=\linewidth]{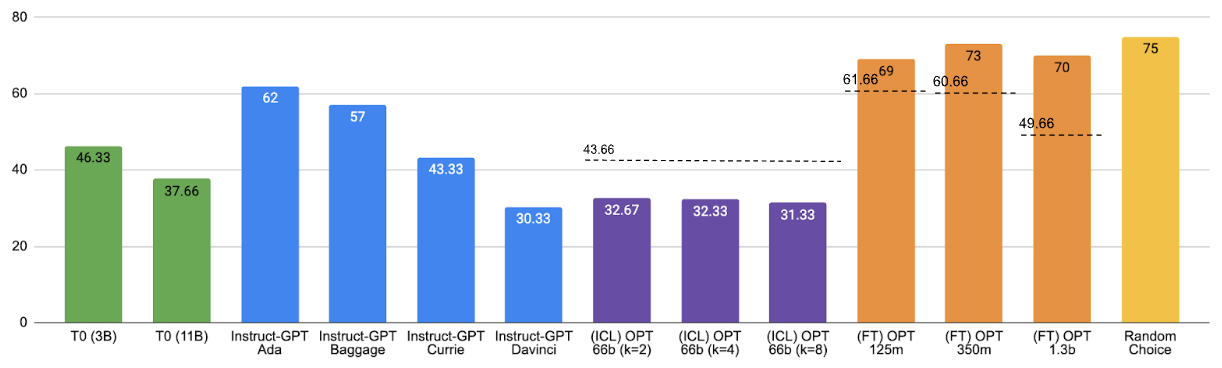}
    \caption{ARC-Easy}
    \label{fig:arc} 
\end{figure} 
\begin{figure}[t!]
    \centering
    \includegraphics[width=\linewidth]{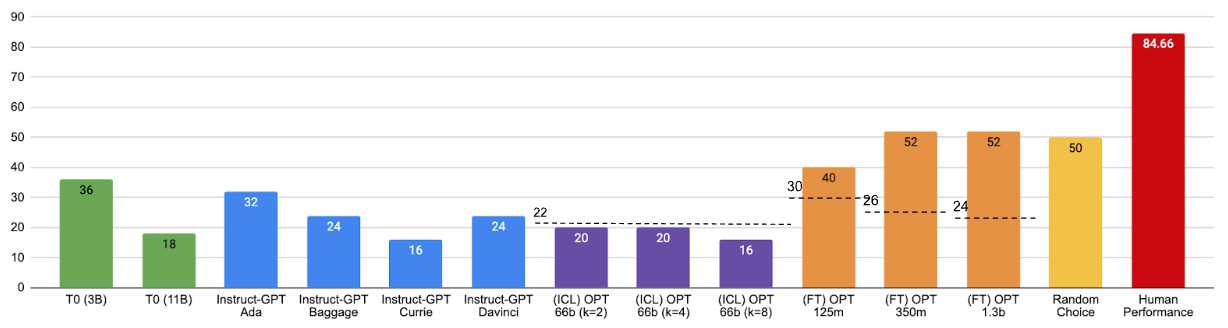}
    \caption{COPA}
    \label{fig:copa} 
\end{figure} 
\begin{figure}[t!]
    \centering
    \includegraphics[width=\linewidth]{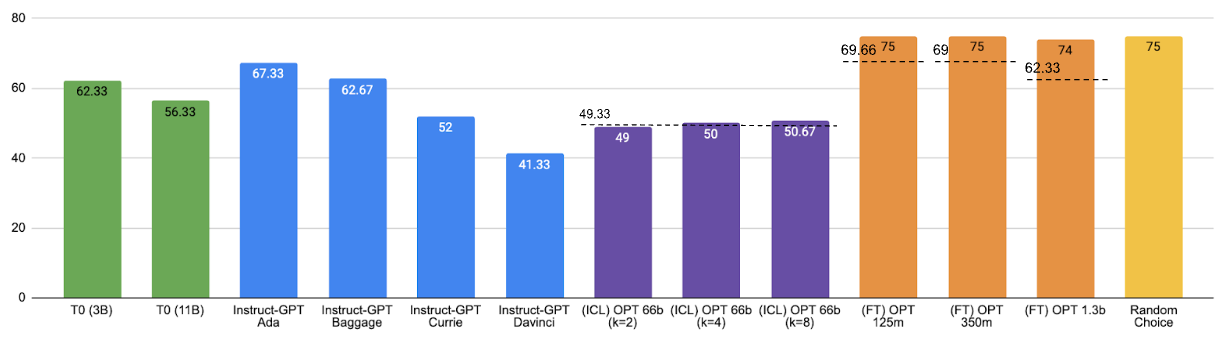}
    \caption{HellaSwag}
    \label{fig:h} 
\end{figure} 
\begin{figure}[t!]
    \centering
    \includegraphics[width=\linewidth]{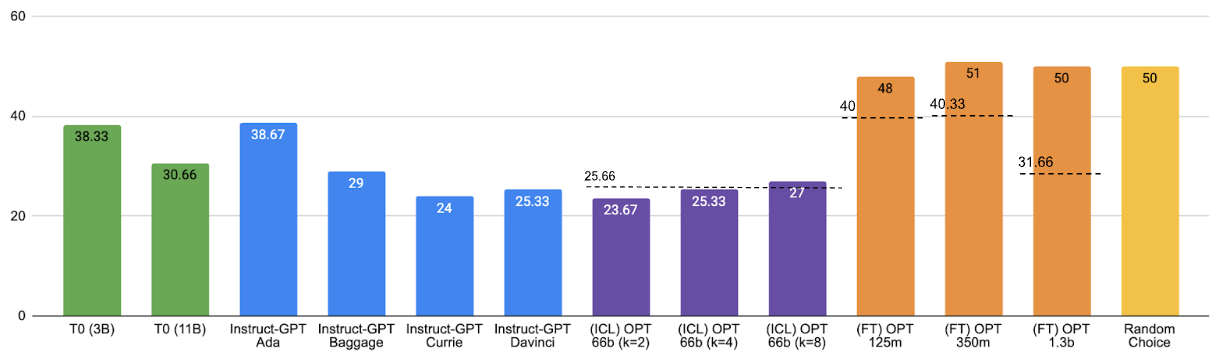}
    \caption{StoryCloze}
    \label{fig:sc} 
\end{figure} 
\begin{figure}[t!]
    \centering
    \includegraphics[width=\linewidth]{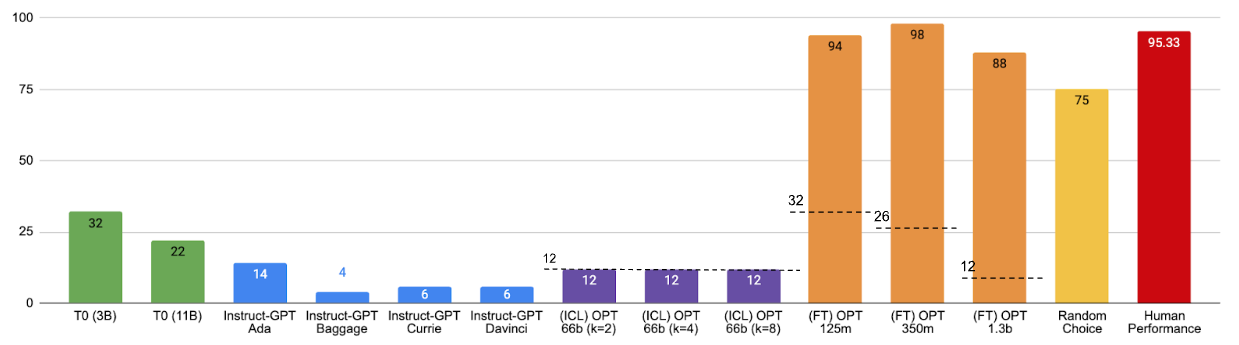}
    \caption{Lambada}
    \label{fig:lambada} 
\end{figure} 
\begin{figure}[t!]
    \centering
    \includegraphics[width=\linewidth]{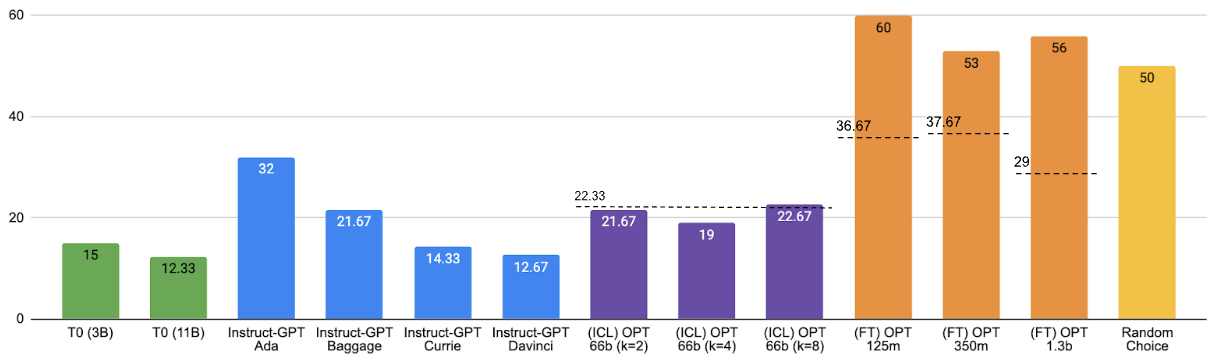}
    \caption{Web Questions}
    \label{fig:wq} 
\end{figure} 
\begin{figure}[t!]
    \centering
    \includegraphics[width=\linewidth]{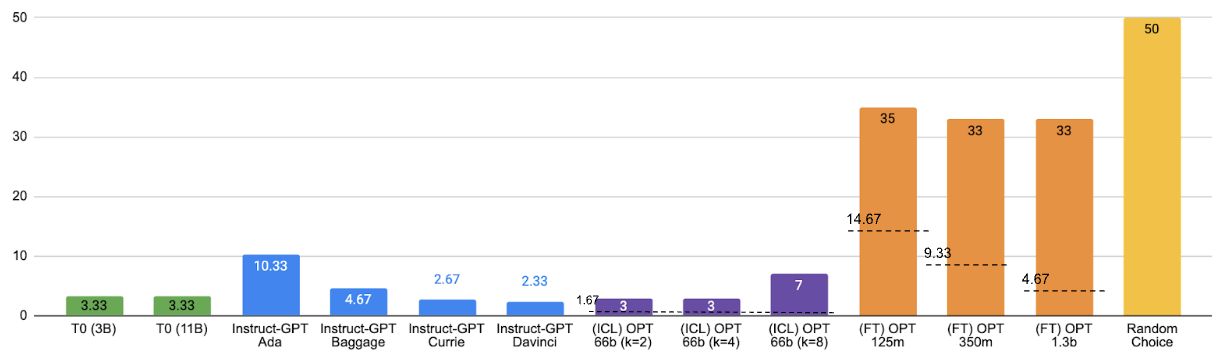}
    \caption{TriviaQA}
    \label{fig:triviaqa} 
\end{figure} 
\begin{figure}[t!]
    \centering
    \includegraphics[width=\linewidth]{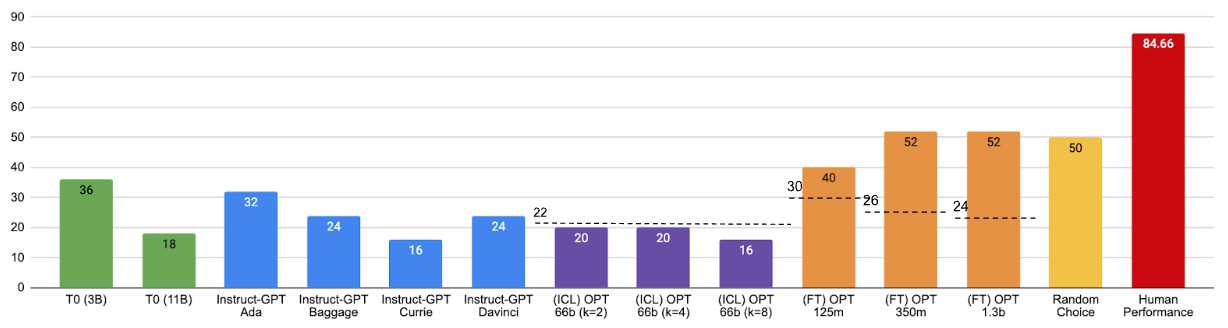}
    \caption{Natural Questions}
    \label{fig:nq} 
\end{figure} 
Figure 5-13 shows the performance of the existing methods on the \textit{negated} prompts. First thing to note is that for all of the tasks, instruction following LMs (T0 and InstructGPT) also show an inverse scaling law where the larger LMs perform worse. Furthermore, while FT seems to help LMs understand \textit{negation}, as shown in Figure \ref{fig:average}, the average score of both original and negated prompts is $\sim$50\%, which means that the mitigation was a result of a trade-off of performance degradation on the original prompts, resulting in a zero-sum game. 

\end{document}